%% file: main.tex
\documentclass[10pt,twocolumn,letterpaper]{article}

\usepackage[pagenumbers]{cvpr} 

\input{preamble}

\definecolor{cvprblue}{rgb}{0.21,0.49,0.74}
\usepackage[pagebackref,breaklinks,colorlinks,allcolors=cvprblue]{hyperref}

\def\paperID{11907} 
\def\confName{CVPR}
\def\confYear{2025}

\title{LesionLocator: Zero-Shot Universal Tumor Segmentation and Tracking\\ in 3D Whole-Body Imaging}

\author{Maximilian Rokuss\textsuperscript{1,2},
Yannick Kirchhoff\textsuperscript{1,2,6},  Seval Akbal\textsuperscript{5}, Balint Kovacs\textsuperscript{1,3}, Saikat Roy\textsuperscript{1,2},\\ Constantin Ulrich\textsuperscript{1,3}, Tassilo Wald\textsuperscript{1,3,4}, Lukas T. Rotkopf\textsuperscript{5}, Heinz-Peter Schlemmer\textsuperscript{5}, Klaus Maier-Hein\textsuperscript{1,3,4,7}\\
\and
\textsuperscript{1}German Cancer Research Center, Division of Medical Image Computing, Germany\\
\textsuperscript{2}Faculty of Mathematics and Computer Science and \textsuperscript{3}Medical Faculty  - Heidelberg University\\
\textsuperscript{4}Helmholtz Imaging, \textsuperscript{5}German Cancer Research Center, Department of Radiology, Germany\\
\textsuperscript{6}HIDSS4Health, Heidelberg
\textsuperscript{7}Pattern Analysis and Learning Group, Heidelberg University Hospital
\\
{\tt\small maximilian.rokuss@dkfz-heidelberg.de}
}

\begin{document}
\maketitle
\input{sec/0_abstract}

\input{sec/1-2_intro}

\input{sec/2_unified_model}
\input{sec/3_evalutation}
\input{sec/5_results}
\input{sec/6_conclusion}
{
    \small
    \bibliographystyle{ieeenat_fullname}
    \bibliography{main}
}

\input{sec/X_suppl}

\end{document}

%% file: preamble.tex
\usepackage{graphicx}
\usepackage{caption}
\usepackage{todonotes}
\usepackage[table,xcdraw,dvipsnames]{xcolor}
\usepackage{booktabs}
\usepackage{multirow}
\usepackage{pgf}
\usepackage{adjustbox}
\usepackage[flushleft]{threeparttable}
\usepackage{pifont}
\newcommand{\cmark}{\ding{51}}%
\newcommand{\xmark}{\ding{55}}%
\usepackage{float}
\usepackage{tablefootnote}

\definecolor{mygreen}{RGB}{128, 255, 0}
\definecolor{myred}{RGB}{229, 20, 0}

\definecolor{goodred}{HTML}{e67c73}
\definecolor{goodyellow}{HTML}{ffd666}
\definecolor{goodgreen}{HTML}{57bb8a}

\pgfmathsetmacro{\minval}{40}  
\pgfmathsetmacro{\maxval}{76}  
\pgfmathsetmacro{\midval}{(\minval+\maxval)/2}  

\newcommand{\colorrange}[1]{%
    \pgfmathsetmacro{\PercentColor}{100.0*(#1-\minval)/(\maxval-\minval)}%
    \ifdim #1 pt < \minval pt
        \cellcolor{goodred}{#1}%
    \else
        \ifdim #1 pt > \maxval pt
            \cellcolor{goodgreen}{#1}%
        \else
            \ifdim #1 pt < \midval pt
                \pgfmathsetmacro{\PercentLow}{200.0*(#1-\minval)/(\maxval-\minval)}%
                \xdef\PercentLow{\PercentLow}%
                \cellcolor{goodyellow!\PercentLow!goodred}{#1}%
            \else
                \pgfmathsetmacro{\PercentHigh}{200.0*(#1-\midval)/(\maxval-\minval)}%
                \xdef\PercentHigh{\PercentHigh}%
                \cellcolor{goodgreen!\PercentHigh!goodyellow}{#1}%
            \fi
        \fi
    \fi
}

\pgfmathsetmacro{\minvala}{50}  
\pgfmathsetmacro{\maxvala}{94}  
\pgfmathsetmacro{\midvala}{(\minvala+\maxvala)/2}  

\newcommand{\colorrangea}[1]{%
    \pgfmathsetmacro{\PercentColor}{100.0*(#1-\minvala)/(\maxvala-\minvala)}%
    \ifdim #1 pt < \minvala pt
        \cellcolor{goodred}{#1}%
    \else
        \ifdim #1 pt > \maxvala pt
            \cellcolor{goodgreen}{#1}%
        \else
            \ifdim #1 pt < \midvala pt
                \pgfmathsetmacro{\PercentLow}{200.0*(#1-\minvala)/(\maxvala-\minvala)}%
                \xdef\PercentLow{\PercentLow}%
                \cellcolor{goodyellow!\PercentLow!goodred}{#1}%
            \else
                \pgfmathsetmacro{\PercentHigh}{200.0*(#1-\midvala)/(\maxvala-\minvala)}%
                \xdef\PercentHigh{\PercentHigh}%
                \cellcolor{goodgreen!\PercentHigh!goodyellow}{#1}%
            \fi
        \fi
    \fi
}

\pgfmathsetmacro{\minvalb}{40}  
\pgfmathsetmacro{\maxvalb}{80}  
\pgfmathsetmacro{\midvalb}{(\minvalb+\maxvalb)/2}  

\newcommand{\colorrangeb}[1]{%
    \pgfmathsetmacro{\PercentColor}{100.0*(#1-\minvalb)/(\maxvalb-\minvalb)}%
    \ifdim #1 pt < \minvalb pt
        \cellcolor{goodred}{#1}%
    \else
        \ifdim #1 pt > \maxvalb pt
            \cellcolor{goodgreen}{#1}%
        \else
            \ifdim #1 pt < \midvalb pt
                \pgfmathsetmacro{\PercentLow}{200.0*(#1-\minvalb)/(\maxvalb-\minvalb)}%
                \xdef\PercentLow{\PercentLow}%
                \cellcolor{goodyellow!\PercentLow!goodred}{#1}%
            \else
                \pgfmathsetmacro{\PercentHigh}{200.0*(#1-\midvalb)/(\maxvalb-\minvalb)}%
                \xdef\PercentHigh{\PercentHigh}%
                \cellcolor{goodgreen!\PercentHigh!goodyellow}{#1}%
            \fi
        \fi
    \fi
}
\pgfmathsetmacro{\minvalc}{40}  
\pgfmathsetmacro{\maxvalc}{81}  
\pgfmathsetmacro{\midvalc}{(\minvalc+\maxvalc)/2}  

\newcommand{\colorrangec}[1]{%
    \pgfmathsetmacro{\PercentColor}{100.0*(#1-\minvalc)/(\maxvalc-\minvalc)}%
    \ifdim #1 pt < \minvalc pt
        \cellcolor{goodred}{#1}%
    \else
        \ifdim #1 pt > \maxvalc pt
            \cellcolor{goodgreen}{#1}%
        \else
            \ifdim #1 pt < \midvalc pt
                \pgfmathsetmacro{\PercentLow}{200.0*(#1-\minvalc)/(\maxvalc-\minvalc)}%
                \xdef\PercentLow{\PercentLow}%
                \cellcolor{goodyellow!\PercentLow!goodred}{#1}%
            \else
                \pgfmathsetmacro{\PercentHigh}{200.0*(#1-\midvalc)/(\maxvalc-\minvalc)}%
                \xdef\PercentHigh{\PercentHigh}%
                \cellcolor{goodgreen!\PercentHigh!goodyellow}{#1}%
            \fi
        \fi
    \fi
}
\pgfmathsetmacro{\minvald}{40}  
\pgfmathsetmacro{\maxvald}{82}  
\pgfmathsetmacro{\midvald}{(\minvald+\maxvald)/2}  

\newcommand{\colorranged}[1]{%
    \pgfmathsetmacro{\PercentColor}{100.0*(#1-\minvald)/(\maxvald-\minvald)}%
    \ifdim #1 pt < \minvald pt
        \cellcolor{goodred}{#1}%
    \else
        \ifdim #1 pt > \maxvald pt
            \cellcolor{goodgreen}{#1}%
        \else
            \ifdim #1 pt < \midvald pt
                \pgfmathsetmacro{\PercentLow}{200.0*(#1-\minvald)/(\maxvald-\minvald)}%
                \xdef\PercentLow{\PercentLow}%
                \cellcolor{goodyellow!\PercentLow!goodred}{#1}%
            \else
                \pgfmathsetmacro{\PercentHigh}{200.0*(#1-\midvald)/(\maxvald-\minvald)}%
                \xdef\PercentHigh{\PercentHigh}%
                \cellcolor{goodgreen!\PercentHigh!goodyellow}{#1}%
            \fi
        \fi
    \fi
}
\pgfmathsetmacro{\minvale}{40}  
\pgfmathsetmacro{\maxvale}{85}  
\pgfmathsetmacro{\midvale}{(\minvale+\maxvale)/2}  

\newcommand{\colorrangee}[1]{%
    \pgfmathsetmacro{\PercentColor}{100.0*(#1-\minvale)/(\maxvale-\minvale)}%
    \ifdim #1 pt < \minvale pt
        \cellcolor{goodred}{#1}%
    \else
        \ifdim #1 pt > \maxvale pt
            \cellcolor{goodgreen}{#1}%
        \else
            \ifdim #1 pt < \midvale pt
                \pgfmathsetmacro{\PercentLow}{200.0*(#1-\minvale)/(\maxvale-\minvale)}%
                \xdef\PercentLow{\PercentLow}%
                \cellcolor{goodyellow!\PercentLow!goodred}{#1}%
            \else
                \pgfmathsetmacro{\PercentHigh}{200.0*(#1-\midvale)/(\maxvale-\minvale)}%
                \xdef\PercentHigh{\PercentHigh}%
                \cellcolor{goodgreen!\PercentHigh!goodyellow}{#1}%
            \fi
        \fi
    \fi
}

%% file: sec/0_abstract.tex
\begin{abstract}
In this work, we present \textit{LesionLocator}, a framework for zero-shot longitudinal lesion tracking and segmentation in 3D medical imaging, establishing the first end-to-end model capable of 4D tracking with dense spatial prompts. Our model leverages an extensive dataset of 23,262 annotated medical scans, as well as synthesized longitudinal data across diverse lesion types. The diversity and scale of our dataset significantly enhances model generalizability to real-world medical imaging challenges and addresses key limitations in longitudinal data availability. LesionLocator outperforms all existing promptable models in lesion segmentation by nearly 10 dice points, reaching human-level performance, and achieves state-of-the-art results in lesion tracking, with superior lesion retrieval and segmentation accuracy. LesionLocator not only sets a new benchmark in universal promptable lesion segmentation and automated longitudinal lesion tracking but also provides the first open-access solution of its kind, releasing our synthetic 4D dataset and model to the community, empowering future advancements in medical imaging. Code is available at: \url{www.github.com/MIC-DKFZ/LesionLocator}
\end{abstract}

%% file: sec/1-2_intro.tex
\section{Introduction}
\label{sec:intro}

\begin{figure}[t]
  \centering
  \includegraphics[width=\linewidth]{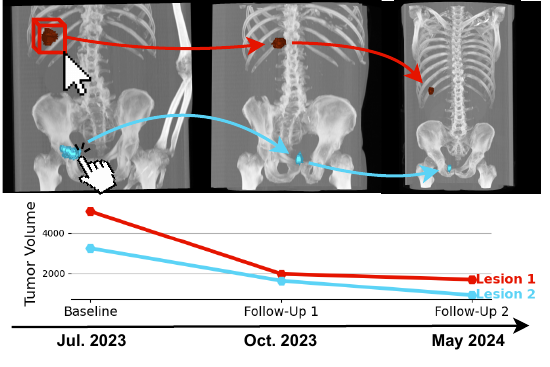}
  \caption{\textbf{Task Overview.} This figure illustrates the core setup and challenges of our task: enabling accurate segmentation and tracking of tumors across multiple follow-up scans from a single initial prompt. This task is inherently difficult due to variability in patient positioning, imaging window, scan protocols, and irregular intervals between imaging sessions. By allowing the user to simply mark a tumor in the initial scan, our approach automates consistent segmentation and tracking across all subsequent scans, streamlining tumor burden assessment and disease progression monitoring.}
  \label{fig:fig0}
\end{figure}

As the annual volume of CT scans continues to rise, and with cancer incidence projected to increase 47\% by 2040, the demand on radiologists continues to intensify \cite{cancer_burden, ct_volume_rise, workload_radiologist}. Furthermore, cancer patients often require multiple imaging exams and monitoring~\cite{recurrent_ct} throughout treatment, placing additional strain on clinical resources. Automated lesion segmentation and tracking have the potential to partially alleviate this workload, supporting increasingly accurate and efficient assessments of tumor burden, disease progression, and radiomics analysis \cite{gillies2016radiomics}. Although recent deep learning models have demonstrated strong performance in segmenting anatomical structures like organs~\cite{bassi2024touchstonebenchmark} using large labeled datasets~\cite{abdomenatlas8k, ji2022amos}, lesion segmentation remains significantly more challenging \cite{intrabench, segvol, rokuss2024fdgpsmahitchhikersguide}. Pathological structures can \textbf{a)} vary widely in appearance, \textbf{b)} arise anywhere in the body, and \textbf{c)} present different degrees of malignancy, making cross-dataset generalization difficult. Developing models that generalize well across diverse lesion types remains an open problem, with efforts like the Universal Lesion Segmentation (ULS) challenge \cite{uls_challenge} beginning to explore solutions through datasets that encompass lesion variability.\\


\noindent Recent breakthroughs in \textit{promptable} foundation models~\cite{wei2022emergent, bommasani2022opportunitiesrisksfoundationmodels}, such as the Segment Anything Model (SAM)~\cite{segment_anything}, have revolutionized computer vision with \textit{zero-shot} segmentation using spatial prompts. These methods show promise for clinical applications, enabling real-time, interactive segmentation. The success of SAM in natural images has driven adaptations to medical imaging, using prompts like points and bounding boxes \cite{roy2023sammd, medsam, cheng2023sammed2d, scribbleprompt}, offering flexibility beyond traditional supervised models constrained by specific training distributions \cite{isensee_nnu-net_2021, swinunetr, mednext, medsegdiff}.

\noindent Currently, the majority of promptable models in medical imaging are designed for 2D images, limiting their ability to capture the complex spatial structures inherent in 3D medical data, which is crucial for clinically relevant tasks such as accurate tumor segmentation \cite{medsam, tang2020clicklesionrecistmeasurement, scribbleprompt}. Although some works have extended prompt-based methods to 3D segmentation, they often rely on multiple prompt types (e.g. class-labeled points, text-annotated boxes) to achieve optimal results, thus restricting their generalizability across diverse tasks \cite{segvol, vista3d}. Other models are specifically trained on a single lesion type or dataset, which limits their flexibility and applicability to new, unseen lesions and body regions~\cite{prism, 3dsamadapter}. 

\noindent Finally, a crucial avenue remains under-explored using concurrent 2D or even 3D promptable methods: \textit{the temporal dimension}. In clinical practice, prior imaging and patient history are vital for accurate diagnosis and monitoring. However, current promptable segmentation methods do not leverage information from previous scans for longitudinal tracking. Available tracking methods are similarly limited: they focus only on tracking point locations without segmentation~\cite{sam_tracking, transformer_lesion_tracker, siemens_tracker}, assume lesions are already segmented across all scans~\cite{santoro2024automated, di2023graph}, or require a multi-step process that decouples segmentation from tracking \cite{hering_tracking}. Inspired by recent advancements that highlight the value of temporal context in natural image segmentation, including SAM 2~\cite{time_does_tell, imagenet_worth_video, sam2}, we introduce the first 4D promptable framework \textit{unifying segmentation and tracking} for medical imaging.\\

\noindent Our proposed model, \textit{\textbf{LesionLocator}}, accepts point or box prompts around lesions, accurately segments them, and tracks them across follow-up scans, thus providing a powerful tool for longitudinal lesion monitoring. The model efficiently propagates prompts over time in an end-to-end trainable framework, leveraging masks from prior scans. To overcome the scarcity of longitudinal datasets -- a significant bottleneck for robust tracking -- we introduce a novel augmentation technique that generates synthetic longitudinal data from single-timepoint images, enabling effective training on large-scale longitudinal data. \textit{LesionLocator} achieves strong generalization across six out-of-distribution zero-shot lesion segmentation tasks, covering diverse lesion types and body regions. It surpasses existing single-timepoint segmentation methods and establishes a new benchmark in volumetric tracking on multi-timepoint data, outperforming all state-of-the-art methods. To promote further research, we will release both the synthetic longitudinal dataset and model weights, creating the first open-source model for lesion tracking. Our main contributions are:

\begin{itemize}
    \item \textbf{Human-Level Performance in Zero-Shot Universal Lesion Segmentation:} Our model achieves unprecedented accuracy in prompt-based 3D lesion segmentation, outpacing competitors by nearly 10 dice points and reaching human inter-rater variability.

    \item \textbf{Unified Segmentation and Tracking Framework:} LesionLocator introduces the first prompt-based 4D framework for both segmentation and tracking, leveraging temporal data to improve longitudinal lesion monitoring.

    \item \textbf{Synthetic Longitudinal Data Generation:} We address the scarcity of public longitudinal datasets with a novel data augmentation method that generates synthetic longitudinal scans from single-timepoint data, which we release to support further research.
\end{itemize}

%% file: sec/2_unified_model.tex
\section{Unifying Segmentation and Tracking}
\label{sec:method}

\begin{figure*}[t]
  \centering
  \includegraphics[width=\textwidth]{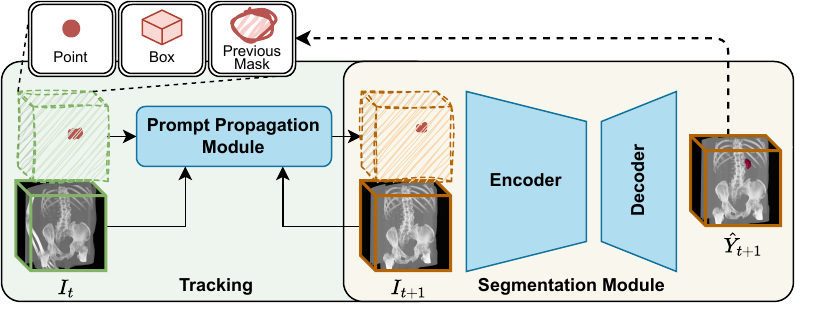}
  \caption{\textbf{Overview of the Proposed Lesion Tracking Pipeline.} Our model receives user-provided prompts on the patient's initial scan, which are then propagated through time by the \textit{Prompt Propagation Module}, enabling accurate lesion tracking across timepoints. Lesion delineation is performed by passing these propagated prompts to the \textit{Segmentation Module}. Crucially, we introduce the propagation of predicted masks from the previous scan as prompt for the current scan, allowing for autoregressive segmentation throughout time series.}
  \label{fig:fig1}
\end{figure*}

We present \textit{LesionLocator}, a promptable framework for versatile lesion segmentation and tracking across longitudinal 3D medical images. Our approach has three core components: (i) \textbf{zero-shot segmentation} of all types of lesions throughout the whole body based on user prompts, (ii) \textbf{prompt propagation} for efficient longitudinal lesion tracking of lesions across sequential scans, and a (iii) \textbf{synthetic longitudinal dataset} to enhance robustness and generalization across timepoints. An overview of our pipeline is shown in Fig. \ref{fig:fig1}.

\subsection{Problem Formulation}

Let $I_t \in \mathbb{R}^{H \times W \times D}$ be a 3D image and $Y_t \in \mathbb{R}^{H \times W \times D}$ be its corresponding segmentation mask at timepoint $t$. If \( p_t \in \{point, box\} \) is a user-provided prompt, our segmentation model \( f_\theta \) predicts the segmentation mask \( \hat{Y}_t \in \mathbb{R}^{H \times W \times D}\):
\begin{equation}
    \hat{Y}_t = f_\theta(I_t, p_t)
\end{equation}

\noindent For a sequence of images $\{I_t, I_{t+1}, ... I_{t+n}\}$ and initial prompt $p_t$, the task of lesion tracking may be expressed as:
\begin{equation}
LL(\{I_t, ..., I_{t+n}\}, p_t) = \{\hat{Y}_t, ...,\hat{Y}_{t+n}\}
\label{eqn:tracking}
\end{equation}

\noindent where $LL$ denotes our \textit{LesionLocator} framework. Notably, during the tracking process, the user provides the prompt solely at timepoint $t$. Therefore, we introduce a \textit{Prompt Propagation Module} \( g \) to transfer this information and locate the structure of interest in the subsequent image. It calculates the deformation field $\Phi_t = g(I_t, I_{t+1})$ from image \( I_{t} \) to \( I_{t+1} \) and warps the prompt accordingly. The propagated prompt is used to generate the segmentation mask for the next timepoint:
\begin{align}
    \hat{Y}_{t+1} &= f_\theta(I_{t+1}, \Phi_t \circ p_{t})
    \label{eqn:prompt_tracking}
\end{align}
Crucially, to leverage shape and size information throughout the temporal dimension, we propose an autoregressive approach, where the generated mask \( \hat{Y}_{t} \) from the previous scan is utilized as an input prompt for the next examination $I_{t+1}$. We may now redefine \cref{eqn:prompt_tracking} as:
\begin{equation}
\hat{Y}_{t+1} = f_\theta(I_{t+1}, \Phi_t \circ \hat{Y}_{t}) \text{~~where~~} \hat{Y}_{0} = f_\theta(I_0, p_0)
\label{eqn:final}
\end{equation}

\noindent We jointly train our segmentation and tracking model end-to-end on consecutive timepoints using a combined segmentation and propagation loss \( \mathcal{L} = \mathcal{L}_{\text{seg}} + \mathcal{L}_{\text{prop}} \). The segmentation loss \( \mathcal{L}_{\text{seg}} = \mathcal{L}_{\text{ce}} + \mathcal{L}_{\text{dice}} \) integrates cross-entropy and soft-Dice loss, while the prompt propagation loss \( \mathcal{L}_{\text{prop}} = \mathcal{L}_{\text{sim}}(I_{t+1}, \Phi \circ I_{t}) + \lambda \mathcal{L}_{\text{reg}}(\Phi) \) includes a normalized cross correlation similarity measure between consecutive images and a GradICON~\cite{gradicon} deformation field regularizer, with \( \lambda \geq 0 \) as a weighting parameter. This joint training strategy allows the segmentation network to adjust to inaccuracies in prompt propagation, thereby enhancing robustness across sequential scans. Furthermore, the propagation network benefits from the additional supervision provided by the segmentation loss. This represents the first end-to-end trainable framework for promptable lesion segmentation and tracking across longitudinal medical images.

\subsection{Network Architecture}

Despite the widespread shift towards Transformer-based models in 2D computer vision, UNet-based backbones~\cite{isensee_nnu-net_2021, mednext} continue to dominate 3D medical image segmentation tasks, as evidenced by recent benchmarks~\cite{bassi2024touchstonebenchmark, nnunet_revisited}, and their frequent success in major 3D medical imaging competitions~\cite{uls_challenge, topcowchallenge, lnq2023challenge, autopet}. Therefore, unlike most existing interactive segmentation models~\cite{segvol, sammed3d, sam2, segment_anything, medsam, cheng2023sammed2d}, we employ a densely promptable 3D UNet. We posit that spatial prompts, such as points or boxes, directly align with a convolutional model’s spatial inductive bias, making them well-suited for input on the highest resolution level of the network. By preserving spatial alignment in the input space, the model can allocate more capacity to segmentation itself, rather than combining prompt and image features in latent space, as common in Transformer-based approaches. Specifically, we employ a residual encoder UNet~\cite{nnunet_revisited}, with prompts as additional channels as our backbone segmentation network. As for the propagation module, we use a multi-resolution, multi-step network comprising several UNets, pretrained on medical images \cite{gradicon,unigradicon}. Details are provided in Appendix \ref{sec:training_appendix}. We train our model on an extensive, large-scale collection of annotated datasets for the segmentation task (Sec. \ref{sec:single_timepoint_training}), and incorporate both real and synthetic multi-timepoint data for lesion tracking (Sec. \ref{sec:tracking_training}).






\subsection{Single-Timepoint Promptable Segmentation}
\label{sec:single_timepoint_training}

\begin{figure}[t]
  \centering
  \includegraphics[width=0.99\linewidth]{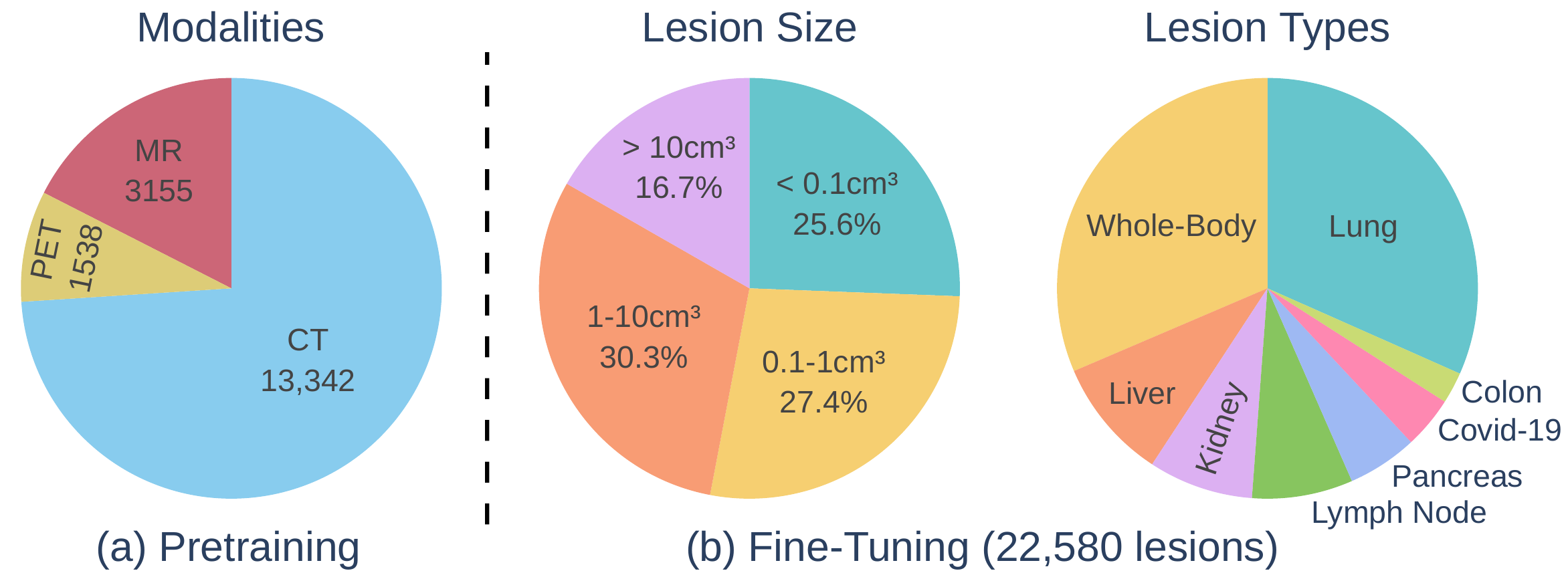}
  \caption{\textbf{Dataset Statistics.} (a) Overview of the pretraining dataset modality composition. (b) Distribution of lesion sizes and types in the fine-tuning dataset.}
  \label{fig:dataset_statistics}
\end{figure}

To build a strong anatomical foundation for lesion segmentation, we adopt a two-stage approach: \textbf{1)} large-scale supervised pretraining on a diverse set of medical imaging data, followed by \textbf{2)} specialized fine-tuning on lesion-specific datasets. Overall, \textit{these combined datasets encompass 4.3 million image slices}, marking a significant advancement in training zero-shot segmentation models.\\

\noindent \textbf{Large-Scale Multi-Modal Pretraining.} We first perform supervised pretraining on a broad collection of 47 publicly available 3D medical segmentation datasets, covering a wide range of anatomical structures and pathological variations (Appendix Tab. \ref{tab:pretraining_datasets}). This dataset collection spans multiple imaging modalities, including CT~\cite{antonelli2021medical, BTCV, lidc, BTCV2, structseg, lambert2019segthor, NHpancreas, verse1, verse2, verse3, ripseg, heller2023kits21, ji2022amos, abdomenatlas8k, totalseg, FLARE22, Luo2023-zk, Jin2021-jo, luo2022word, Ma-2021-AbdomenCT-1K, qu2023abdomenatlas, Podobnik2023-mp, topcowchallenge, Rister2019-tm}, different Magnetic Resonance Imaging (MRI) sequences~\cite{Antonelli2022, isles2015, Litjens2014, Bernard2018, Carass2017, CHAOSdata2019, Campello2021, prostatex, Muslim2022-qt, Grovik2020-un, Shapey2021-iz, Liew2018-wy, ji2022amos, bratsgli1, bratsgli2, bratsgli3} and Positron Emission Tomography (PET)~\cite{autopet, Andrearczyk2023-ii} shown in Fig \ref{fig:dataset_statistics}a. This comprehensive pretraining phase on 18,035 3D images annotated with various structures, follows the MultiTalent~\cite{multitalent} framework for multi-dataset training, to allow holistic anatomical understanding across modalities. \\

\noindent \textbf{Fine-Tuning on Lesion-Specific Data.} Following pretraining, we fine-tune the segmentation model on a targeted collection of datasets featuring lesions, comprising 5,227 images with a total of 22,580 annotated lesions, detailed in Fig \ref{fig:dataset_statistics}b. These lesions are derived from a plethora of publicly available datasets~\cite{deeplesion, covid19_junma, heller2023kits21, FLARE23, lidc, lndb, antonelli2021medical, Antonelli2022, nih_lymph, NSCLC-Radiomics, autopet, covid19_challenge} outlined in Appendix Tab. \ref{tab:finetuning_datasets}.\\

\subsection{Enabling Lesion Tracking at Scale}
\label{sec:tracking_training}


We aim to train a \textit{single model} for zero-shot volumetric lesion tracking over time, capable of scaling across multiple datasets and diverse disease characteristics. To achieve this, we propose to train both prompt propagation and segmentation module jointly on \textit{longitudinal} imaging data. However, this poses a non-trivial challenge due to the limited availability of radiological image-based time series datasets. Training deep networks to model the diversity of disease progression over time relies heavily on access to real-world patient data, which remains a critical bottleneck. To address this, we \textbf{a)} annotate lesions in a real dataset of medical imaging time series, and \textbf{b)} propose a novel data augmentation pipeline that synthesizes longitudinal data from single-timepoint images.\\

\noindent \textbf{Real Data: Annotating Longitudinal Image Series.} An intra-institutional longitudinal dataset was annotated by a radiology specialist consisting of patients diagnosed with malignant melanoma, a type of cancer known for producing various tumors throughout the body. Each of the 60 patients in the dataset has at least two scans at different timepoints, with individual lesions consistently labeled across time to enable accurate tracking, amounting to a total number of 159 scans with five lesions on average.\\

\noindent \textbf{Synthetic Data: Simulating Disease Progression via \textit{Anatomy-Informed} Transformations.} The usage of our real dataset is necessary but insufficient for effective longitudinal modeling of lesions at scale. To further address the scarcity of public longitudinal datasets, we generate synthetic longitudinal data to augment the training set. Recent advancements in synthetic dataset generation have demonstrated the efficacy of instance-level augmentations in improving dataset diversity \cite{dataset_enhancement_instance_level_augmentations}. Building on this concept, we simulate random lesion growth and shrinkage to model disease progression by adapting the anatomy-informed transformation approach \cite{kovacs2023anatomy}, which defines the deformation field $V$ around each lesion calculated as the gradient of a Gaussian kernel $G_{\sigma s}$ convolved with the indicator function $S_{lesion}$ and multiplied with an amplitude $A$:
\begin{equation}
    V = \nabla (G_{\sigma s} * S_{lesion}(x,y,z)) \cdot A(x, y, z).
\end{equation}

\noindent To model random progression, we modulate the amplitude $A$ by assigning a random scalar to each voxel separately within a specified range $r(x, y, z) \sim \mathcal{U}(r_{\text{min}}, r_{\text{max}})$ and smooth this field by using a Gaussian kernel $G_{\sigma r}$:
\begin{equation}
    A(x, y, z) = A \cdot (G_{\sigma r} * r(x, y, z)).
\end{equation}

\begin{figure}[t]
  \centering
  \includegraphics[width=1.0\linewidth]{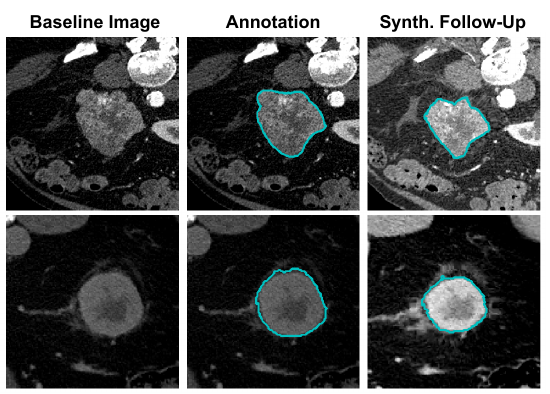}
  \caption{Examples from the synthetic dataset used to augment the training process for lesion tracking. The dataset simulates disease progression through random lesion progression, based on anatomy-informed transformations and image augmentations. Additional examples are provided in the Appendix.}
  \label{fig:syn_dataset}
\end{figure}

\noindent Owing to the substantial variability in lesion size compared to organs, applying fixed transformation parameters may either destroy smaller lesions or yield minimal changes in larger ones. To mitigate this, we implement a multi-stage transformation approach, stacking smaller amplitude transformations to achieve more realistic outcomes. These size and shape alterations of lesions are further combined with image-level intensity and spatial augmentations that mimic variations in examination conditions, enabling the synthetic generation of multi-timepoint datasets from single-timepoint images.\\
After filtering our large lesion-specific training dataset (Fig. \ref{fig:dataset_statistics}b) for suitable images based on sufficient scan size, we compiled a dataset of 2,728 scans, for which we synthesized an additional timepoint. Exemplary images are shown in Fig. \ref{fig:syn_dataset}. We pretrain our tracking model on the synthetic data and subsequently fine-tune with real longitudinal scans.

\subsection{Prompt Simulation}

During training, we randomly select images and then sample a random instance from the ground truth mask using connected components or available instance annotations to simulate various types of prompts.

\noindent \textbf{Point.} A point prompt is generated by selecting a random pixel within the foreground. To enhance this area, specifically for localized convolutional kernels, we apply a ball-shaped footprint with a radius of 5 pixels around the click.

\noindent \textbf{Bounding Boxes.} We generate a bounding box by calculating the minimum ROI that encloses the label, and expand each dimension by \(r \sim U[0, 10]\) pixels to account for variability in human annotations.

\noindent \textbf{Prior Mask.} For tracking across longitudinal images, the model is trained with inputs that include sampled points or bounding boxes in the above fashion or ground truth segmentation masks propagated from the previous timepoint to the current image. Note, that we only use the ground truth masks during training and propagate generated masks from earlier prompts during inference.

%% file: sec/3_evalutation.tex
\section{Evaluation}
\label{sec:eval}

We benchmark LesionLocator against state-of-the-art promptable segmentation models tailored for biomedical imaging and tumor segmentation, as well as existing volumetric lesion tracking frameworks.\\

\noindent \textbf{Single-Timepoint Zero-Shot Evaluation.} To rigorously validate our segmentation model, we curated six held-out downstream datasets (Appendix Tab. \ref{tab:test_datasets}), each featuring tumor lesions of various types and locations throughout the body. Specifically, the test set encompasses lung nodules~\cite{rider_lung}, colorectal~\cite{colorectal_liver} and primary liver cancer\cite{hcc_tace_seg}, adrenal tumor~\cite{adrenal_tumor}, whole-body melanoma lesions as well as malignant lymph nodes~\cite{lnq2023challenge}. This setup simulates an out-of-distribution clinical scenario, accounting for variations in patient age, race, gender, and scanner type. The model is prompted with either points or bounding boxes, and notably, neither our model nor the baseline models have seen these datasets during training.\\

\begin{table*}[ht!]
\centering
\setlength{\tabcolsep}{4pt} 
\begin{adjustbox}{width=\textwidth}
\begin{tabular}{llccccccc|c}
\toprule
Dim & Model & Prompt & \shortstack{Colorectal\\ Liver Tumor} & \shortstack{Adrenal\\Tumor} & \shortstack{Primary\\Liver Cancer} & \shortstack{Lymph Node\\Metastases} & \shortstack{Lung\\Tumor}	 & \shortstack{Whole-body\\Melanoma}  & \shortstack{Avg.\\Dice}\\
\midrule
\multirow{4}{*}{2D} & SAM~\cite{segment_anything} & Box & \colorrange{58.77} / 55.88 & \colorrangea{58.96} / 26.24 & \colorrangeb{58.10} / 24.80 & \colorrangec{51.39} / 46.09 & \colorranged{50.55} / 43.06 & \colorrangee{55.22} / 48.51 & 55.50\\
                    & SAM-Med2D~\cite{cheng2023sammed2d} & Box & \colorrange{41.54} / 39.91 & \colorrangea{52.72} / 26.38 & \colorrangeb{55.01} / 24.51 & \colorrangec{43.46} / 38.38 & \colorranged{43.26} / 37.32 & \colorrangee{40.18} / 33.05 & 46.03\\
                    & MedSAM~\cite{medsam} & Box & \colorrange{54.38} / 56.77 & \colorrangea{68.63} / 32.92 & \colorrangeb{65.00} / 31.47 & \colorrangec{52.04} / 52.08 & \colorranged{59.37} / 53.62 & \colorrangee{53.31} / 51.26 & 53.42\\
                    & ScribblePrompt~\cite{scribbleprompt} & Box & \colorrange{55.25} / 60.79 & \colorrangea{68.56} / 44.24 & \colorrangeb{56.15} / 34.93 & \colorrangec{55.91} / 61.48 & \colorranged{63.25} / 63.85 & \colorrangee{67.88} / 68.65 & 61.17\\
                    \cline{1-3}
2D+\textit{t}       & SAM2~\cite{sam2}      & Box   & \colorrange{69.94} / 62.23 & \colorrangea{75.31} / 33.29 & \colorrangeb{66.25} / 29.29 & \colorrangec{70.02} / 60.86 & \colorranged{65.08} / 53.02 & \colorrangee{70.00} / 58.65 & 69.10\\
                    \cline{1-3}
\multirow{7}{*}{3D} & SAM-Med3D~\cite{sammed3d} & Point & \colorrange{48.32} / 40.39 & \colorrangea{82.40} / 55.91 & \colorrangeb{63.18} / 27.77 & \colorrangec{19.81} / 14.93 & \colorranged{39.77} / 26.62 & \colorrangee{39.08} / 29.98 & 48.76\\
                    & NVIDIA VISTA\cite{vista3d}& Point+Class & - & \colorrangea{46.22} / 27.71 & \colorrangeb{61.95} / 33.30 & - & \colorranged{51.18} / 45.91 & \colorrangee{27.10} / 22.20 & - \\
                    & SegVol~\cite{segvol} & Point+Text & \colorrange{68.28} / 66.10 & \colorrangea{83.59} / 55.85 & \colorrangeb{71.83} / 40.93 & \colorrangec{51.09} / 42.51 & \colorranged{68.94} / 63.53 & \colorrangee{58.39} / 49.45 & 67.02\\
                    & SegVol~\cite{segvol} & Box+Text & \colorrange{58.35} / 57.63 & \colorrangea{90.52} / 68.86 & \colorrangeb{75.01} / 45.36 & \colorrangec{71.85} / 67.18 & \colorranged{73.23} / 71.41 & \colorrangee{58.62} / 48.91 & 71.26\\
                    & ULS Model~\cite{uls_challenge} & Point & \colorrange{68.28} / 69.36 & \colorrangea{82.97} / 58.94 & \colorrangeb{65.16} / 37.39 & \colorrangec{74.95} / 72.47 & \colorranged{76.53} / 72.71 & \colorrangee{77.64} / 77.48 & 74.25\\ \addlinespace  \cline{2-10} \addlinespace
                    
                    & \multirow{2}{*}{\textit{\textbf{Ours}}} & Point & \colorrange{75.38} / 78.15 & \colorrangea{89.10} / 67.68 & \colorrangeb{78.39} / 52.86 & \colorrangec{75.60} / 73.82 & \colorranged{77.18} / 73.20 & \colorrangee{82.58} / 83.93 & \textbf{79.71}\\
                    &   & Box & \colorrange{74.05} / 77.85 & \colorrangea{92.04} / 76.42 & \colorrangeb{85.71} / 60.10 & \colorrangec{80.63} / 81.29 & \colorranged{82.51} / 82.60 & \colorrangee{84.66} / 87.89 & \textbf{83.26}\\
\midrule
 \multicolumn{3}{r}{Human Dice Inter-Rater Variability} & 76 \cite{liver_interrater} & - & 84 \cite{liver_interrater} & 80 \cite{lymph_interrater} & 81-85 \cite{lung_interrater} & 80-85\cite{hering2024improving}\\
\bottomrule
\end{tabular}
\end{adjustbox}
\caption{\textbf{Zero-Shot Evaluation of Promptable Models on Lesion Segmentation.} We present Dice/NSD metrics for state-of-the-art 2D and 3D models across six held out lesion datasets, with color-coded scores normalized per column for clear comparison. Human inter-observer variability of studies from the respective lesion type is provided as an upper bound reference. 2D models used slice-wise bounding box prompts, while 3D models employed both point and box prompts (+ class). The results highlight the superior performance of 3D models, with our method using box prompts achieving the highest scores. VISTA results are excluded for certain datasets due to training overlap or unavailable class labels. Due to poor performance, results from 2D point-based prompts are omitted.}
\label{tab:results_st}
\end{table*}

\noindent \textbf{Promptable Segmentation Baselines.} We compare our model against several state-of-the-art 2D and 3D zero-shot interactive segmentation methods, particularly those tailored for biomedical imaging and lesion segmentation. Comparisons include SAM \cite{segment_anything}, which was originally trained on natural images, and its medical variants: SAM-Med2D \cite{cheng2023sammed2d}, trained on biomedical images of various modalities, MedSAM \cite{medsam}, trained with box prompts on 1.5M medical segmentations, and ScribblePrompt \cite{scribbleprompt}, trained on 65 diverse medical datasets covering both healthy anatomy and lesions. For \textbf{2D models}, we compute slice-wise segmentations prompting a 3D bounding box. Additionally, we evaluate SAM2 \cite{sam2}, a video-based segmentation model treated in a 2D+\textit{t} manner by considering image slices as sequential frames. For \textbf{3D models}, we benchmark against SAM-Med3D \cite{sammed3d}, a 3D adaptation of SAM, and other transformer-based models like NVIDIA VISTA \cite{vista3d} and SegVol \cite{segvol}, trained on diverse medical image datasets containing both healthy structures and lesions. These models are tested using 3D bounding boxes or center-point prompts. We also include the click-based lesion segmentation model from the ULS challenge \cite{uls_challenge}.

\noindent \textbf{Lesion Tracking Baselines.} Existing lesion tracking methods are often constrained by (1) retrieving only point locations in follow-up images without segmentation~\cite{sam_tracking, transformer_lesion_tracker, siemens_tracker}, (2) assuming all lesions are pre-identified and segmented across scans~\cite{santoro2024automated, di2023graph}, or (3) employing multi-step processes that separate segmentation from tracking~\cite{hering_tracking}. In this work, we focus on methods that address both retrieval and segmentation. Hering et al.~\cite{hering_tracking} exemplify a multi-step approach by training nnUNet~\cite{isensee_nnu-net_2021} on 100mm lesion-centered crops for segmentation, followed by classical image registration for tracking. As no public code is available, we reimplemented the registration using Elastix~\cite{simple_elastix} and retrained the nnUNet on our dataset. We also benchmark the ULS model~\cite{uls_challenge} paired with classical deformable image registration, allowing for a comprehensive comparison across tracking methods. We evaluate all tracking models on our real medical time series data using 5-fold cross-validation.\\

\noindent \textbf{Metrics.} For single-timepoint segmentation, we evaluate performance using the Dice Similarity Coefficient (DSC) for mask overlap and the Normalized Surface Dice (NSD) with a 2mm tolerance for boundary alignment. In the tracking setting, prompts are provided on the previous timepoint image, and we assess follow-up segmentations using the following metrics, consistent with related work~\cite{hering_tracking}: Center Point Matching (CPM@25), which calculates the percentage of predicted and ground truth lesion center points within 25mm; Dice@25, representing the Dice score for correctly matched lesions; the Mean Euclidean Distance (MED) between predicted and ground truth lesion centers; and the overall Dice score across all tracked lesions. All metrics are averaged per patient, with further details in Appendix \ref{sec:metrics_appendix}.

%% file: sec/5_results.tex
\section{Results and Discussion}
\label{sec:results}


We first compare our model’s zero-shot segmentation performance against state-of-the-art promptable segmentation models tailored for biomedical imaging and tumor segmentation. We then benchmark our unified tracking model against existing volumetric lesion tracking frameworks and evaluate temporal consistency.

\subsection{Lesion Segmentation at Single-Timepoints}

\textbf{3D Spatial Context enables Enhanced Segmentation.} LesionLocator is trained with full 3D volumes and drastically outperforms \textit{all existing 2D as well as 3D} promptable foundational baselines evaluated for lesion segmentation. In general, Tab. \ref{tab:results_st} shows that recent 3D point and box-prompted models achieve significantly higher average Dice scores than 2D models, except for SAM-Med3D or VISTA. 
In fact, most 3D models, even those prompted only with point prompts, outperform ScribblePrompt, the top-performing 2D model, with box prompts. Notably, adding spatial context in the 2D+t SAM2 model enhances its performance by 15 points over the original 2D SAM, placing it above all other 2D models. However, it still lags behind the performance achieved by fully 3D models. Point-prompted 2D models were excluded due to poor performance. These results underscore the importance of full 3D spatial context in training a generalist method.\\



\noindent \textbf{Superior Zero-Shot Segmentation across Tumor Types.} We demonstrate state-of-the-art segmentation performance across all evaluated lesion types in the held-out test set, spanning colorectal, adrenal, liver, lymph node, lung and whole-body lesions, establishing a new benchmark for universal promtable tumor segmentation. Notably, our model achieves Dice scores of 84.66 on melanoma and 82.51 on lung tumors (3D box), surpassing the next-best ULS Model by a substantial margin of 7 and 6 points, respectively. This trend is consistent across all other lesion types as well, with a remarkable 10.7 Dice-point lead over the closest competitor on liver tumors, highlighting the versatility and robustness of the model.\\

\begin{table*}[h]
\centering
\setlength{\tabcolsep}{4pt} 
\begin{tabular}{lccccc}
\toprule
\textbf{Model} & \textbf{Propagated Prompt} & \textbf{Dice}$\uparrow$ & \textbf{CPM@25}$\uparrow$ & \textbf{Dice@25}$\uparrow$ & \textbf{MED}$\downarrow$\\
\midrule
Yan et.al~\cite{sam_tracking} \textit{(point tracker)} & Point & -- & --* & -- & 6.92 \\
Hering et. al.\cite{hering_tracking} & Point & 58.66 & 82.50 & 69.27 & \underline{4.45} \\
Reg. + ULS Model\cite{uls_challenge} & Point & 55.69 & 79.81 & 67.55 & 5.81 \\
\textbf{\textit{Ours (LesionLocator)}} & Point & \underline{62.62} & \underline{83.62} & \underline{74.45} & 5.09 \\
\midrule
\textbf{Ours} + Synth. Data & Point & 64.33 & 85.40 & 74.61 & 3.89 \\
\textbf{Ours} + Synth. Data & Box & 66.49 & 83.95 & 77.98 & 3.25 \\
\textbf{Ours} + Synth. Data & Prev. Gen. Seg & \textbf{68.31} & \textbf{85.96} & \textbf{79.02} & \textbf{3.12} \\


\bottomrule
\end{tabular}
\caption{\textbf{Lesion Tracking Performance Comparison.} This table reports 5-fold cross-validation results of lesion tracking models on the whole-body melanoma dataset. Metrics include Dice, CPM@25 (lesion matching accuracy), Dice@25 (retrieved lesion segmentation quality), and MED (center point error). As no box-promptable tracking approach currently exists, we compare our model against baseline methods propagating point locations in the upper half. \underline{Underline} indicates the best performance for models using point prompts trained solely on real data. \textbf{Bold} highlights the best overall performance incorporating synthetic data and three different propagated prompt types. \textit{Note:} * indicates not comparable, as segmentation-based methods generating false masks receive a score of zero on empty follow-ups.}
\label{tab:results_tracking}
\end{table*}

\begin{table}[ht]
\centering
\begin{tabular}{lccc}
\toprule
Prompt & Pretraining & Dice$\uparrow$ & NSD$\uparrow$\\
\midrule
point as single pixel & \cmark & 66.56 & 59.12 \\
point as ball region & \cmark & 79.71 & 71.61 \\
\midrule
box & \xmark & 82.44 & 76.37 \\
box & \cmark & 83.26 & 77.69 \\
\bottomrule
\end{tabular}
\caption{Segmentation performance ablation showing the effect of different prompt types and pretraining. The upper two lines compare point prompts implemented as a single pixel versus a spherical region as proposed. The lower two lines ablate the impact of pretraining when using box prompts.}
\label{tab:results_seg_ablation}
\end{table}


\noindent \textbf{LesionLocator Reaches Human-Level Accuracy.} Our model demonstrates remarkable alignment with human inter-rater variability~\cite{lung_interrater,lymph_interrater,liver_interrater,hering2024improving}, an essential indicator for clinical applicability. By aligning with the natural variability found in expert annotations, it offers significant potential for real-world deployment. For instance, in lung tumor segmentation, our model achieves a Dice score of 82.51, closely matching the human variability range of 81-85. Similar results can be observed for liver, lymph and melanoma lesion segmentation. This consistency with human variability underscores the model's potential for providing accurate, clinically viable segmentation.\\

\noindent \textbf{Prompt Design and Pretraining Matters.} The ablation results in Tab. \ref{tab:results_seg_ablation}  highlight the significant effects of prompt design and pretraining on segmentation performance. Using a spherical ball-shaped region for point prompts instead of a single pixel boosts Dice and NSD scores by 13.15 and 12.49 points, respectively. This performance gain likely arises from the limited receptive field of $3\times3\times3$ convolution kernels, limiting the prompt information to a small local neighborhood.
We also ablate the effect of pretraining our segmentation model on the diverse set of annotated medical images which increases Dice and NSD scores by 0.82 and 1.32 respectively. These findings underscore the importance incorporating pretraining for lesion segmentation.

\subsection{Longitudinal Lesion Tracking}


\textbf{State-of-the-art click-based Tracking.} Our model achieves superior performance in click-based lesion tracking, surpassing current methods in their limited scope of using only point prompts without mask propagation and syntetic data, as shown in the upper half of Table \ref{tab:results_tracking}. Tested with only propagating point prompts to the next scan, our method already achieves the highest lesion matching accuracy (CPM@25) of 83.62 and corresponding Dice@25 of 74.45, demonstrating superior segmentation and lesion-tracking precision. While its mean Euclidean distance (MED) of 5.09 in this setup is slightly above that of Hering et al., it still outperforms the ULS model, affirming comparable localization performance. These findings underscore our model's leading performance across key metrics, even within the constraints of propagating only point locations.\\

\noindent \textbf{Synthetic Data enhances Lesion Retrieval.} We successfully address the challenge of longitudinal data scarcity by our proposed method of incorporating synthetic time-series data, thus leading to substantial improvements as shown in the second part of Tab. \ref{tab:results_tracking}. Our model trained exclusively on melanoma data using point prompts, already demonstrates SOTA performance in Dice, CPM@25 and Dice@25, but underperforms in MED, which is an indicator of accurate lesion retrieval. However, when our synthetic data is incorporated, MED drops \textit{significantly} to 3.8 and lesion matching improves, as seen by the increase in CPM@25 to 85.40. This indicates a higher recall of lesions in the subsequent image, thereby \textit{raising the overall Dice score by almost 2 points while still using point prompts}.\\


\noindent \textbf{Mask Propagation yields Optimal Results.} While the added size prior of box prompts is seen to be better than point prompts, our method of propagating previous segmentation masks shows the best performance as it leverages prior shape information. This is indicated by top results across all metrics, confirming the value of propagating prior segmentation information for more accurate lesion tracking. LesionLocator achieves a retrieval rate of 86\%, with a corresponding Dice score of 79, and the average distance between the centers of retrieved and ground truth lesions is just 3mm. Notably, we use the mask from a \textit{point-prompted} segmentation in the initial image.\\

\begin{figure}[t]
  \centering
  \includegraphics[width=1.0\linewidth]{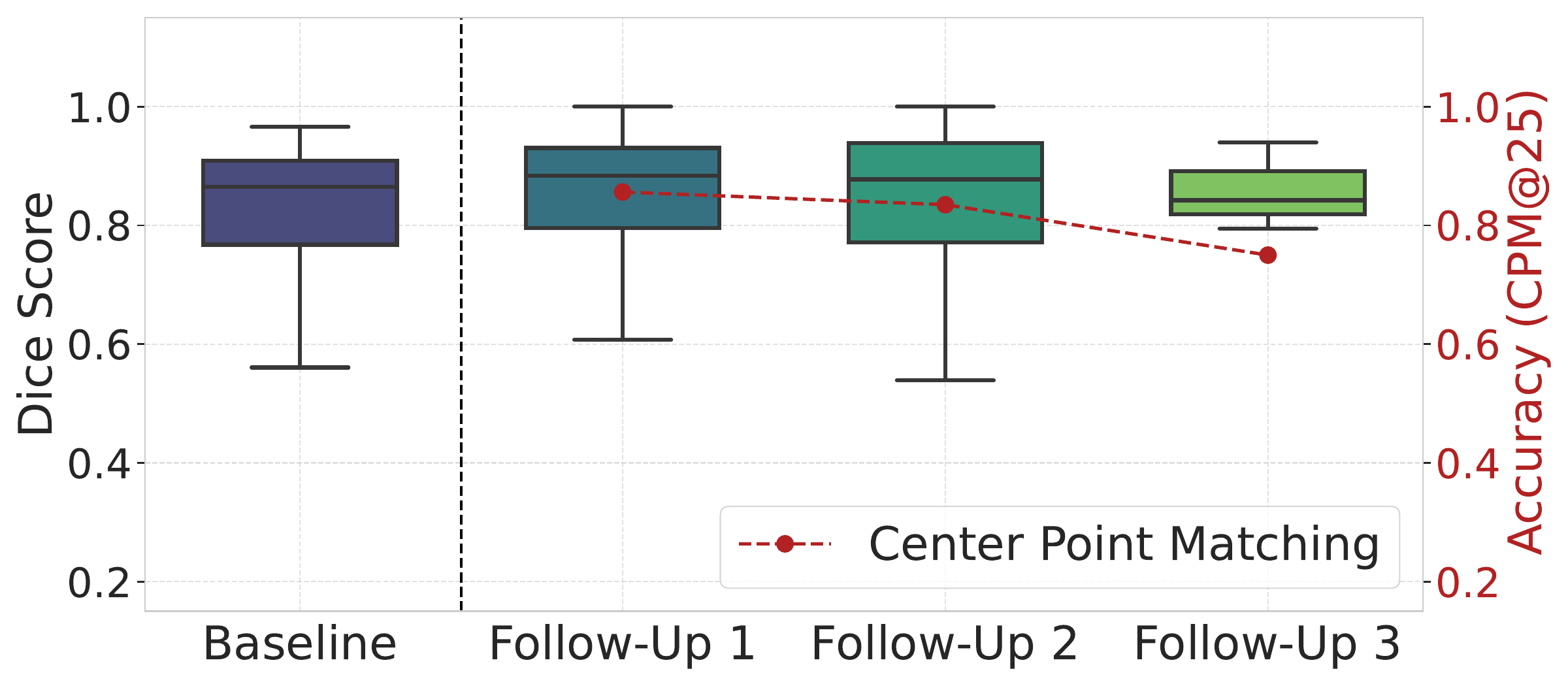}
  \caption{
\textbf{Consistent Lesion Tracking Performance Over Time.}\\
For the baseline scan the initial Dice score distribution is shown using the LesionLocator segmentation model with box prompts. For follow-up scans, tracking is performed autoregressively, as proposed, using prior masks as prompts. Tracking accuracy relative to the baseline is measured as CPM@25 (lesion matches within 25mm) with corresponding Dice@25 (Dice score of matched lesions). The Dice for matched lesions remains consistently high, with matching accuracy above 80\% and only a slight decrease over time. \underline{Note:} Only a single patient in our real longitudinal dataset has a Follow-Up 3 scan, so this distribution is based on one scan with 4 lesions, of which 3 were correctly matched.}
  \label{fig:tracking_autoreg}
\end{figure}


\noindent \textbf{Consistently High Tracking Performance.} LesionLocator demonstrates exceptional temporal tracking consistency, as shown in Fig. \ref{fig:tracking_autoreg}. The initial Dice distribution on the prompted baseline scan highlights the strong performance of our segmentation module. The bar plots of subsequent timepoints showcase consistently high Dice scores achieved through our autoregressive mask prompts, even when trained solely on consecutive images pairs. This underscores the generalizability of our longitudinal training approach. With minimal performance degradation over time, LesionLocator excels in lesion matching (CPM@25), ensuring reliable, sustained tracking across multiple scans.\\

\begin{figure}[H]
\raggedright
  \scriptsize\textbf{\;\;\; Ground Truth \;\;\;\;\; Hering et al~\cite{hering_tracking} \;\;\;\;\; Reg. ULS~\cite{uls_challenge} \;\;\;\;\;\;\;\;\;\;\;\;\;\; \footnotesize\textit{Ours}}\par\smallskip
  \centering
  \includegraphics[width=1.0\linewidth]{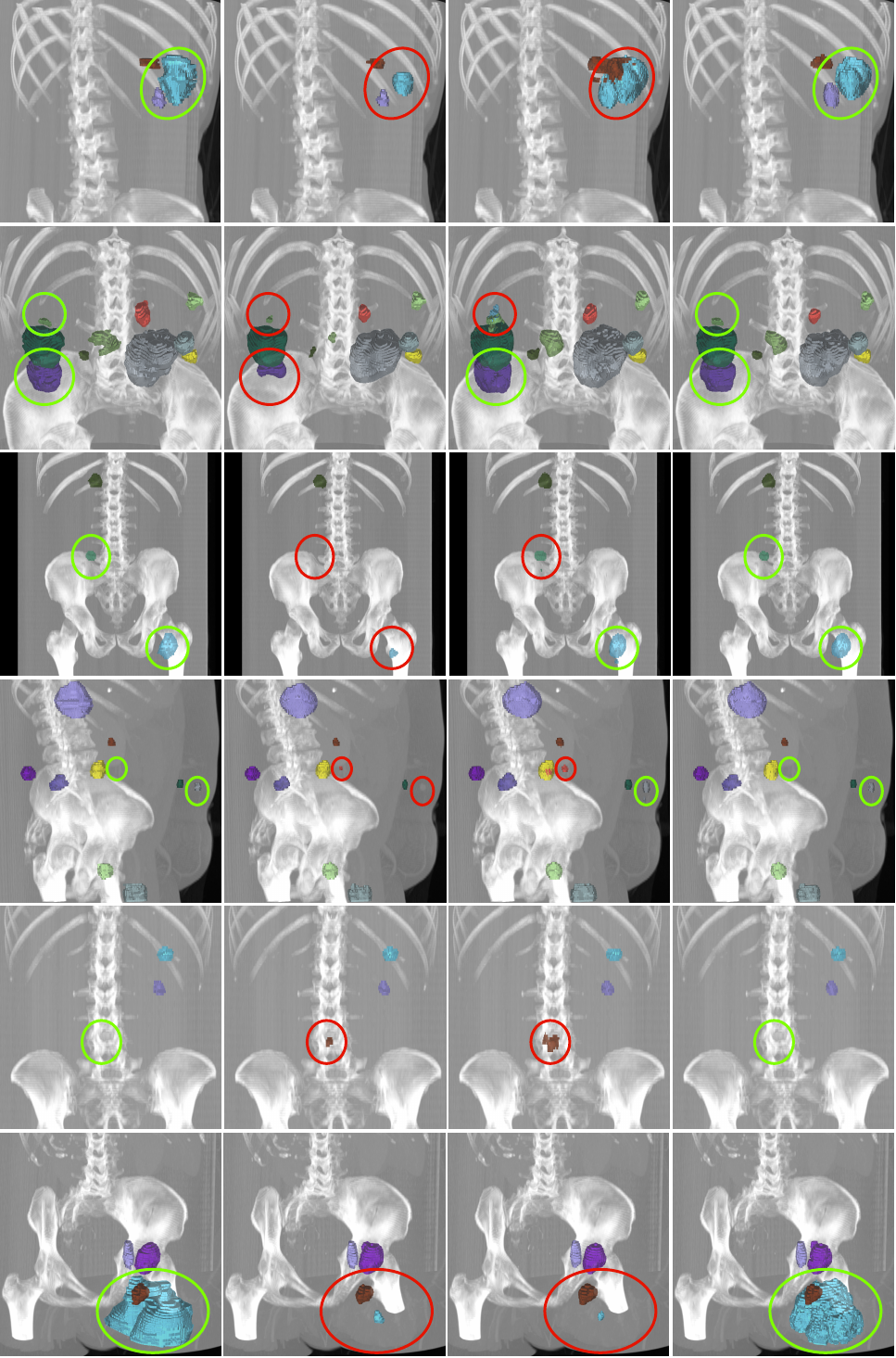}
  \caption{
\textbf{Qualitative Results on Follow-Up.} \textcolor{mygreen}{Green} circles indicate correctly matched and segmented lesions. \textcolor{myred}{Red} circles denote failed or wrong retrieval and poor segmentation performance.}
  \label{fig:qualitative}
\end{figure}

\noindent \textbf{Qualitative Evaluation.} Figure \ref{fig:qualitative} presents tracking results on the melanoma dataset, demonstrating LesionLocator’s superior ability to accurately match and segment lesions over time. In contrast, competing models struggle with lesion retrieval, failing to correctly identify vanishing lesions or suffer from mismatches and degraded segmentation performance.\\



\noindent \textbf{Limitations \& Future Work.} While our approach is demonstrated primarily on CT, it is inherently modality-independent; pretrained across PET, CT, and MRI, it can be fine-tuned for other imaging types only limited by dataset availability. Moreover, accurate distinction of closely located or merging lesions remains a challenge, however, re-identification with an additional click can enable further reliable tracking. Additionally, our work also opens up future

\noindent possibilities of replacing the point prompt on the initial scan by a manual segmentation, thus allowing their integration into the lesion tracking process.


%% file: sec/6_conclusion.tex
\section{Conclusion}
\label{sec:conclusion}

This paper introduces a novel framework for medical imaging that integrates 3D promptable zero-shot segmentation and longitudinal lesion tracking, addressing critical gaps in current methods. The proposed synthetic longitudinal data augmentation technique overcomes the significant challenge of limited multi-timepoint datasets. By leveraging both real and synthetic data, our model not only achieves state-of-the-art performance in segmenting diverse lesion types but also excels in tracking lesions across multiple timepoints, making it a valuable tool for clinical monitoring with promising implications for real-world clinical applications.

\section*{Acknowledgments}

The present contribution is supported by the Helmholtz Association under the
joint research school "HIDSS4Health – Helmholtz Information and Data Science
School for Health". This work was partly funded by Helmholtz Imaging (HI),
a platform of the Helmholtz Incubator on Information and Data Science.

%% file: sec/X_suppl.tex
\clearpage
\setcounter{page}{1}
\maketitlesupplementary
\captionsetup{width=\textwidth}

\section*{Overview}

This document provides supplementary details on the methods and experiments presented in the main paper:\\

\begin{itemize}
    \item \textbf{Training} (Section~\ref{sec:training_appendix}): Details the architecture and parameters for both pretraining and fine-tuning of the segmentation model, followed by a discussion of the longitudinal tracking setup, including prompt propagation for multi-timepoint analysis.

    \item \textbf{Synthetic Data Generation} (Section~\ref{sec:syn_data_appendix}): Outlines the lesion- and image-level augmentations used to simulate realistic disease progression and imaging variations, supporting robust training across diverse longitudinal patterns.

    \item \textbf{Promptable Segmentation Baselines} (Section~\ref{sec:promt_baselines_appendix}): Describes the state-of-the-art promptable models benchmarked against our approach, specifically those adapted for medical imaging and segmentation.

    \item \textbf{Evaluation Metrics} (Section~\ref{sec:metrics_appendix}): Explains the metrics used for segmentation and tracking accuracy.

    \item \textbf{Additional Experimental Results} (Section~\ref{sec:results_appendix}): Provides further comparisons against supervised baselines on downstream tasks.
\end{itemize}

\section{Training}
\label{sec:training_appendix}

\subsection{Segmentation Model}

\textbf{Pretraining.} The pretraining pipeline was implemented using the nnU-Net framework \cite{isensee_nnu-net_2021}, specifically utilizing a ResEncL U-Net architecture \cite{extendingnnunet,revisited}. The model was trained for 4,000 epochs with a patch size of [192,192,192] and a batch size of 24. All images were resampled to a cubic 1mm resolution and z-score normalized. Training was performed with an initial learning rate of 1e-2, employing polynomial learning rate decay and the SGD optimizer. Following the MultiTalent strategy \cite{multitalent}, datasets were sampled inversely proportional to the square root of the number of images per dataset, ensuring balanced training across datasets. A summary of the pretraining datasets can be found in Table \ref{tab:pretraining_datasets}.\

\noindent \textbf{Fine-Tuning.} Fine-tuning of the promptable segmentation model was conducted using the combined lesion datasets outlined in Table \ref{tab:finetuning_datasets}. During data loading, images were first randomly picked, followed by random sampling of lesion instances to ensure diverse training samples. The pretrained weights were used to initialize the main body of the network, while the stem and head were randomly initialized. Fine-tuning was carried out with a reduced initial learning rate of 1e-3. Prompts were input directly at the first level of the network, concatenated with the 3D image volume. To accommodate higher-resolution images, we employed an axial spacing of 0.8mm, resulting in an overall spacing of 0.8x0.8x1mm. The patch size was accordingly adjusted to [224,224,160], and CT images were normalized following the nnU-Net protocol. The model was trained for 2,000 epochs with a batch size of 3.

\begin{table*}[t]
\centering
\caption{\textbf{Overview over all datasets used for the supervised pretraining.} This table provides a comprehensive overview of the datasets utilized for the supervised pretraining of our model. It includes a total of 47 datasets, detailing the name of each dataset, the number of images, the imaging modality employed, the specific anatomical targets, and links for direct access. These diverse datasets cover a wide range of anatomical structures and pathological conditions, ensuring a robust foundation for subsequent lesion segmentation tasks.}
\label{tab:pretraining_datasets}
\resizebox{\textwidth}{!}{ 
\begin{tabular}{lccrr}
\toprule
Name & Images & Modality & Target & Link \\
\midrule
Decatlon Task 2 \cite{antonelli2021medical,simpson2019large}& 20 & MRI & Heart & \url{http://medicaldecathlon.com/} \\
Decatlon Task 3 \cite{antonelli2021medical,simpson2019large}& 131 & CT & Liver, L. Tumor & \url{http://medicaldecathlon.com/} \\
Decatlon Task 4 \cite{antonelli2021medical,simpson2019large}& 208 & MRI & Hippocampus & \url{http://medicaldecathlon.com/} \\
Decatlon Task 5 \cite{antonelli2021medical,simpson2019large}& 32 & MRI & Prostate & \url{http://medicaldecathlon.com/} \\
Decatlon Task 6 \cite{antonelli2021medical,simpson2019large}& 63 & CT & Lung Lesion & \url{http://medicaldecathlon.com/} \\
Decatlon Task 7 \cite{antonelli2021medical,simpson2019large}& 281 & CT & Pancreas, P. Tumor & \url{http://medicaldecathlon.com/} \\
Decatlon Task 8 \cite{antonelli2021medical,simpson2019large}& 303 & CT & Hepatic Vessel, H. Tumor & \url{http://medicaldecathlon.com/} \\
Decatlon Task 9 \cite{antonelli2021medical,simpson2019large}& 41 & CT & Spleen & \url{http://medicaldecathlon.com/} \\
Decatlon Task 10 \cite{antonelli2021medical,simpson2019large}& 126 & CT & Colon Tumor & \url{http://medicaldecathlon.com/} \\
ISLES2015 \cite{isles2015}& 28 & MRI & Stroke Lesion & \url{http://www.isles-challenge.org/ISLES2015/} \\
BTCV \cite{BTCV}& 30 & CT & 13 abdominal organs & \url{https://www.synapse.org/Synapse:syn3193805/wiki/89480} \\
LIDC \cite{lidc}& 1010 & CT & Lung lesion & \url{https://www.cancerimagingarchive.net/collection/lidc-idri/} \\
Promise12 \cite{Litjens2014}& 50 & MRI & Prostate & \url{https://zenodo.org/records/8026660} \\
ACDC \cite{Bernard2018}& 200 & MRI & RV cavity, myocardium, LV cavity & \url{https://www.creatis.insa-lyon.fr/Challenge/acdc/databases.html} \\
ISBILesion2015 \cite{Carass2017}& 42 & MRI & MS Lesion & \url{https://iacl.ece.jhu.edu/index.php/MSChallenge} \\
CHAOS \cite{CHAOSdata2019}& 60 & MRI & Liver, Kidney (L\&R), Spleen & \url{https://zenodo.org/records/3431873} \\
BTCV 2 \cite{BTCV2}& 63 & CT & 9 abdominal organs & \url{https://zenodo.org/records/1169361\#.YiDLFnXMJFE} \\
StructSeg Task1 \cite{structseg} & 50 & CT & 22 OAR Head \& neck & \url{https://structseg2019.grand-challenge.org} \\
StructSeg Task2 \cite{structseg} & 50 & CT & Nasopharynx cancer & \url{https://structseg2019.grand-challenge.org/Home/} \\
StructSeg Task3 \cite{structseg} & 50 & CT & 6 OAR Lung & \url{https://structseg2019.grand-challenge.org/Home/} \\
StructSeg Task4 \cite{structseg} & 50 & CT & Lung Cancer & \url{https://structseg2019.grand-challenge.org/Home/} \\
SegTHOR \cite{lambert2019segthor} & 40 & CT & heart, aorta, trachea, esophagus & \url{https://competitions.codalab.org/competitions/21145} \\
NIH-Pan \cite{clark_cancer_2013,NHpancreas}  & 82 & CT & Pancreas & \url{https://wiki.cancerimagingarchive.net/display/Public/Pancreas-CT} \\
VerSe2020 \cite{verse1,verse2,verse3} & 113 & CT & 28 Vertebrae & \url{https://github.com/anjany/verse} \\
M\&Ms \cite{Campello2021,10103611}& 300 & MRI & l. ventricle, r. ventricle, l. ventri. myocardium & \url{https://www.ub.edu/mnms/} \\
ProstateX \cite{prostatex} & 140 & MRI & Prostate lesion & \url{https://www.aapm.org/GrandChallenge/PROSTATEx-2/} \\
RibSeg \cite{ripseg}& 370 & CT & Rips & \url{https://github.com/M3DV/RibSeg?tab=readme-ov-file} \\
MSLesion \cite{Muslim2022-qt} & 48 & MRI & MS Lesion & \url{https://data.mendeley.com/datasets/8bctsm8jz7/1} \\
BrainMetShare \cite{Grovik2020-un} & 84 & MRI & Brain Metastases & \url{https://aimi.stanford.edu/brainmetshare} \\
CrossModa22 \cite{Shapey2021-iz}& 168 & MRI & vestibular schwannoma, cochlea & \url{https://crossmoda2022.grand-challenge.org/} \\
Atlas22 \cite{Liew2018-wy}& 524 & MRI & stroke lesion & \url{https://atlas.grand-challenge.org/} \\
KiTs23 \cite{heller2023kits21} & 489 & CT & Kidneys, k. Tumors, Cysts & \url{https://kits-challenge.org/kits23/} \\
AutoPet2 \cite{Gatidis2022-ms}& 1014 & PET,CT & Lesions & \url{https://autopet-ii.grand-challenge.org/} \\
AMOS \cite{ji2022amos}& 360 & CT,MRI & 15 abdominal organs & \url{https://amos22.grand-challenge.org/} \\
BraTs23 \cite{Karargyris2023,bratsgli1,bratsgli2,bratsgli3} & 1251 & MRI & Glioblastoma & \url{https://www.synapse.org/Synapse:syn51156910/wiki/621282} \\
AbdomenAtlas1.0 \cite{li2024well,qu2023abdomenatlas}& 5195 & CT & 8 abdominal organs & \url{https://github.com/MrGiovanni/AbdomenAtlas?tab=readme-ov-file} \\
TotalSegmentatorV2 \cite{totalseg}& 1180 & CT & 117 classes of whole body & \url{https://github.com/wasserth/TotalSegmentator} \\
Hecktor2022 \cite{Andrearczyk2023-ii}& 524 & PET,CT & nodal Gross Tumor Volumes (Head\&Neck) & \url{https://hecktor.grand-challenge.org/} \\
FLARE \cite{FLARE22} & 50 & CT & 13 abdominal organs & \url{https://flare22.grand-challenge.org/} \\
SegRap \cite{Luo2023-zk}& 120 & CT & 45 OARs (Head\&Neck) & \url{https://segrap2023.grand-challenge.org/} \\
SegA \cite{Radl2022-fz,Jin2021-jo,Pepe2020-fh}& 56 & CT & Aorta & \url{https://multicenteraorta.grand-challenge.org/data/} \\
WORD \cite{luo2022word,liao2023comprehensive}& 120 & CT & 16 abdominal organs & \url{https://github.com/HiLab-git/WORD} \\
AbdomenCT1K \cite{Ma-2021-AbdomenCT-1K} & 996 & CT & Liver, Kidney, Spleen, pancreas & \url{https://github.com/JunMa11/AbdomenCT-1K} \\
DAP-ATLAS \cite{jaus2023towards}& 533 & CT & 142 classes of whole body & \url{https://github.com/alexanderjaus/AtlasDataset} \\
CTORG \cite{Rister2019-tm} & 140 & CT & lung, brain, bones, liver, kidneys and bladder & \url{https://www.cancerimagingarchive.net/collection/ct-org/} \\
HanSeg \cite{Podobnik2023-mp}& 42 & CT & OAR (Head\&Neck) & \url{https://han-seg2023.grand-challenge.org/} \\
TopCow \cite{topcowchallenge}& 200 & CT,MRI & vessel components of CoW & \url{https://topcow23.grand-challenge.org/} \\
\bottomrule
\end{tabular}
}
\end{table*}

\begin{table*}[!ht]
    \centering
    \caption{\textbf{Fine-Tuning Datasets for Lesion Segmentation.}  This table summarizes the 16 datasets used for fine-tuning our promptable lesion segmentation model. Each dataset contributes a collection of annotated images targeting various types of lesions. For each dataset, we provide the name, the number of images, the specific types of lesions targeted, and links to access the datasets for further exploration.}
    \label{tab:finetuning_datasets}
    \resizebox{\textwidth}{!}{ 
    \begin{tabular}{lclr}
        \toprule
        Name & Images & Target & Link \\
        \midrule
        Deep Lesion & 1093 & Various kinds of lesions & \url{https://nihcc.app.box.com/v/DeepLesion} \\ 
        COVID-19 CT Lung & 10 & Covid -19 & \url{https://zenodo.org/records/3757476} \\ 
        FLARE23 Test Set & 50 & Various kinds of lesions & \url{https://codalab.lisn.upsaclay.fr/competitions/12239} \\ 
        KiTS & 488 & Kidney Lesions & \url{https://kits-challenge.org/kits23/} \\ 
        LIDC & 1010 & Lung Lesions & \url{https://www.cancerimagingarchive.net/collection/lidc-idri/} \\ 
        LNDb & 229 & Lymph nodes & \url{https://lndb.grand-challenge.org} \\ 
        MSD Colon & 126 & Colon Lesions & \url{http://medicaldecathlon.com/} \\ 
        MSD Hepatic Vessels & 303 & Liver Lesions & \url{http://medicaldecathlon.com/} \\ 
        MSD Liver & 118 & Liver Lesions & \url{http://medicaldecathlon.com/} \\ 
        MSD Lung & 63 & Lung Lesions & \url{http://medicaldecathlon.com/} \\ 
        MSD Pancreas & 281 & Pancreas Lesions & \url{http://medicaldecathlon.com/} \\ 
        NIH Lymph & 176 & Lymph nodes & \url{https://www.cancerimagingarchive.net/collection/ct-lymph-nodes/} \\ 
        NSCLC Pleural effusion & 78 & Pleural effusion & \url{https://www.cancerimagingarchive.net/analysis-result/plethora/} \\ 
        NSCLC Radiomics & 503 & Lung Lesions & \url{https://www.cancerimagingarchive.net/collection/nsclc-radiomics/} \\ 
        autoPET & 500 & Melanoma & \url{https://autopet-ii.grand-challenge.org/} \\ 
        COVID-19-20 & 199 & Covid-19 & \url{https://covid-segmentation.grand-challenge.org/COVID-19-20/} \\ 
        \bottomrule
    \end{tabular}
    }
\end{table*}

\subsection{Longitudinal Tracking}

The comprehensive tracking model integrates the single timepoint segmentation network, trained in the above fashion, with the prompt propagation module. We utilize the GradICON~\cite{gradicon} framework as a backbone for the propagation module, initializing it with pretrained weights obtained from a diverse set of image registration datasets~\cite{unigradicon}. Images from both timepoints are resampled to a uniform shape of [175, 175,175] before being fed into the prompt propagation network. We then train both prompt propagation module and segmentation model jointly on the real longitudinal data or first on the synthetic followed by fine-tuning on the real data.\\ 

\noindent Specifically, the propagation module generates a deformation field \(\Phi\), which facilitates the propagation of prompts—these may include points, bounding boxes, or segmentation masks produced by the segmentation model. During training, the propagation network is provided with downsampled versions of the baseline and follow-up images, whereas the segmentation network operates on a higher-resolution cropped region centered around the propagated prompt. The center of this region of interest (ROI) is defined by a random voxel of the propagated prompt during training, while the prompt’s center is used during inference. The ROI  matches the segmentation network’s patch size, and is extracted from the high-resolution image (0.8 x 0.8 x 1 mm) and subsequently processed by the segmentation network to generate the output mask.

\noindent We utilized PyTorch 2.3.1 and conduct experiments on NVIDIA A100 GPUs with 40GB of memory.

\section{Synthetic Data Generation}
\label{sec:syn_data_appendix}

We generate synthetic longitudinal time series data by applying instance-level lesion augmentations in combination with image-level spatial and intensity transformations to single-timepoint images.\\

\subsection{Lesion-Level Augmentations}

To simulate random disease progression, we adapt the anatomy-informed transformation approach \cite{kovacs2023anatomy} to model lesion growth and shrinkage, creating realistic synthetic longitudinal data. Specifically, we construct deformation fields \( V \) around each lesion by computing the gradient of a Gaussian kernel \( G_{\sigma s} \) convolved with a lesion indicator function \( S_{lesion} \), i.e. the lesion ground truth mask, scaled by an amplitude \( A \):

\[
V = \nabla (G_{\sigma s} * S_{lesion}(x,y,z)) \cdot A(x, y, z).
\]

\noindent To introduce variability in progression, we modulate the amplitude \( A \) with a location-dependent random field \( r(x, y, z) \sim \mathcal{U}(r_{\text{min}}, r_{\text{max}}) \), then smooth this field using a Gaussian kernel \( G_{\sigma r} \):

\[
A(x, y, z) = A \cdot (G_{\sigma r} * r(x, y, z)).
\]

\noindent Given the significant size variability of lesions compared to surrounding anatomical structures, fixed transformation parameters can distort smaller lesions or inadequately alter larger ones. To address this, we employ a multi-stage approach, applying a sequence of moderate transformations with parameters adapted to lesion size:\\

\begin{itemize}
    \item  $\mathbf{G_{\sigma s}}$: Size of the Gaussian kernel used to blur the lesion segmentation, adapted based on lesion size, with values ranging from $4-5.5$.

    \item $\mathbf{A}$: Initial amplitude for lesion dilation, randomly sampled across the entire image from the set \([-22, -18, 15, 25]\). Negative values simulate lesion shrinkage, while positive values induce growth.
    
    \item $\mathbf{r(x, y, z)}$: Random voxel-wise scaling field for amplitude \( A \), with values drawn from the range \((-3.5, 3.5)\). This ensures spatial variability, so that the lesion grows or shrinks non-uniformly across 3D space.
    
    \item $\mathbf{G_{\sigma r}}$: Gaussian kernel size, fixed at 3, used to smooth the random modulation field \( r(x, y, z) \).
\end{itemize}

\noindent This transformation approach ensures robust, size-sensitive adjustments to lesions, creating realistic variations in lesion shape and size over time.

\subsection{Image-Level Augmentations}

We further enhance these lesion-level alterations with image-level intensity and spatial augmentations to simulate real-world examination variability. In our augmentation pipeline, we employ the \texttt{batchgeneratorsv2}~\cite{isensee2020batchgenerators} package to streamline the application of spatial and intensity transformations. Below, we detail each transformation included in the pipeline:\\

\begin{itemize}
    \item \textbf{Spatial Transform}:
    We utilize elastic deformations, rotations, scaling, and translations to introduce realistic spatial variations. Key parameters include:
    \begin{itemize}
        \item \textbf{Elastic Deformations}: Applied with a probability of 1.0 to simulate structural variability, with \texttt{elastic\_deform\_scale} and \texttt{elastic\_deform\_magnitude} set to \( (0.05, 0.05) \).
        \item \textbf{Rotation}: Random rotations in the range \((-5^\circ, 5^\circ)\) are applied with a probability of 1.0, introducing slight angular variations.
        \item \textbf{Scaling}: Applied with a probability of 0.5, using scaling factors drawn from \( (0.95, 1.05) \), and set to synchronize across all axes for uniform scaling.
        \item \textbf{Translation}: Minor translations within the range \((-5, 5)\) pixels are applied with a probability of 1.0 to emulate slight spatial shifts in image positioning.
    \end{itemize}

    \item \textbf{Gaussian Noise Transform}:
    We add Gaussian noise to simulate varying noise levels across imaging sessions. Noise variance is sampled from \( (0, 0.05) \) and applied independently across channels, with a probability of 1.0.

    \item \textbf{Gaussian Blur Transform}:
    Gaussian blurring with a sigma range of \( (0.1, 0.2) \) is applied with a probability of 0.1 to replicate the effects of lower scan quality or minor out-of-focus regions. This transform is applied in an unsynchronized manner across channels and axes to maintain realistic variability.

    \item \textbf{Multiplicative Brightness Transform}:
    Brightness adjustments are applied with a probability of 0.15 to emulate diverse lighting conditions, with brightness multipliers drawn from the range \( (0.75, 1.25) \).

    \item \textbf{Contrast Transform}:
    Contrast is adjusted with a probability of 0.15 to simulate different imaging conditions. Contrast levels are sampled from the range \( (0.75, 1.25) \) and applied while preserving the original intensity range to prevent artifacts.

\end{itemize}

\begin{figure*}[t]
  \centering
  \includegraphics[width=0.6\linewidth]{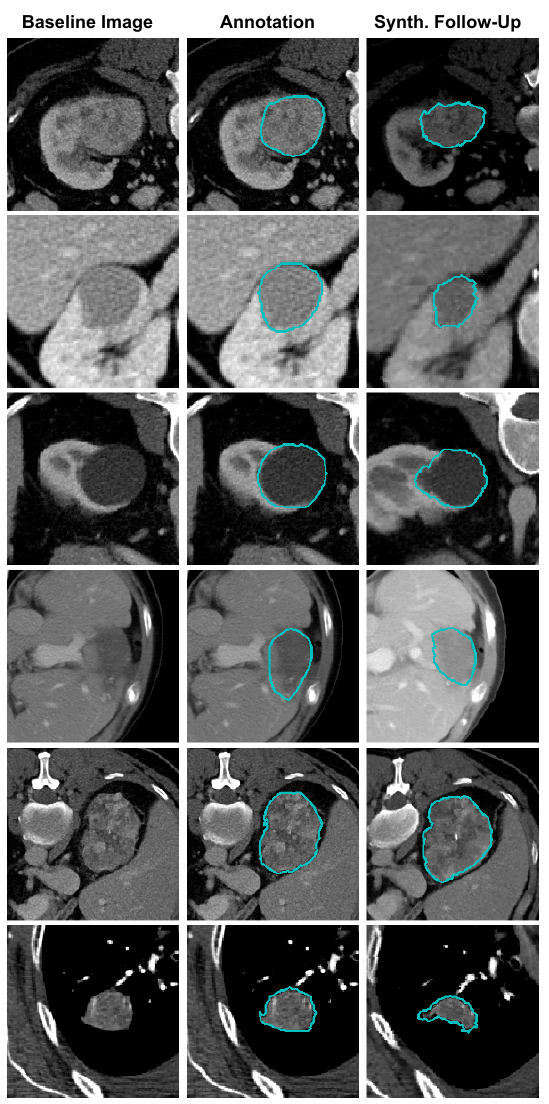}
  \caption{More examples from the synthetic dataset used to augment the training process for lesion tracking. The dataset simulates disease progression through random lesion progression, based on anatomy-informed transformations and image augmentations.}
  \label{fig:syn_dataset_appendix}
\end{figure*}

\noindent We provide additional examples of synthetically generated longitudinal images in Fig. \ref{fig:syn_dataset_appendix}.

\section{Promptable Segmentation Baselines}
\label{sec:promt_baselines_appendix}

The \textbf{Segment Anything Model} (SAM) by META is a leading model from the natural image domain that has inspired numerous researchers to adapt it for radiological medical imaging. While it was trained on 1 billion masks and 11 million images, it did not focus explicitly on radiological data. SAM was the first to popularize interactive segmentation approaches \cite{segment_anything}. \\

\noindent \textbf{MedSAM} is a tailored adaptation of SAM, fine-tuned on 1,570,263 image-mask pairs specifically from the medical domain. Unlike its predecessors, MedSAM is limited to box prompts \cite{medsam}.\\

\noindent \textbf{SAM-Med2D} is a SAM ViT-b model with additional
adapter layers in the image encoder. It was fine-tuned on 4.6 million images and 19.7 million masks from the medical domain using boxes and clicks.\cite{cheng2023sammed2d}.\\

\noindent \textbf{ScribblePrompt} is a state-of-the-art medical segmentation model that supports prompting via points, boxes, or scribbles. It offers the flexibility of utilizing either a UNet or SAM model (ViT-b) backbone; we opted for the UNet backbone due to its superior performance in the performed user study. This model is trained on a comprehensive dataset of 65 diverse medical sources, encompassing a wide range of both healthy anatomical structures and various lesions \cite{scribbleprompt}.\\

\noindent \textbf{SAM2} extends SAM, enhancing its capabilities by incorporating support for video data and increasing the training dataset size \citep{sam2}. In our experiments, we utilize the SAM2.1 Hiera Base Plus checkpoint and evaluate it in a 2D+t configuration, treating axial slices as individual images and interpreting the z-dimension as the temporal axis.\\

\noindent \textbf{SAM-Med3D} introduces a transformer-based 3D image encoder, 3D prompt encoder, and 3D mask decoder. The original model was trained from scratch on 22,000 3D images and 143,000 corresponding 3D masks. \textbf{SAM-Med 3D Turbo} is an enhanced version of SAM-Med 3D, trained on a more extensive dataset collection consisting of 44 datasets for improved performance, which we use in our comparisons. It supports both point and mask prompts \cite{sammed3d}.\\

\noindent \textbf{NVIDIA VISTA} is a 3D segmentation model that supports point prompts in conjunction with class prompts for 127 common human anatomical structures and various lesion types. The model leverages SegResNet~\cite{segresnet} as its backbone CNN, enhanced by SAM’s prompt encoder. It was trained on a comprehensive dataset comprising 11,454 CT volumes, both private and public, which include real and pseudo labels \cite{vista3d}.\\

\noindent \textbf{SegVol} is an interactive 3D segmentation model that utilizes a 3D adaptation of the Vision Transformer (ViT) architecture. It was initially trained on 96,000 unlabelled CT images and subsequently fine-tuned using 6,000 labeled CT images. SegVol supports both point and bounding box prompts as spatial inputs, as well as corresponding text prompts that describe the class. In our experiments, we prompt with "lesion" or "tumor" which increased results \cite{segvol}.\\

\noindent \textbf{ULS model}: Developed for the Universal Lesion Segmentation challenge \cite{uls_challenge}, the ULS model is specifically tailored for lesion segmentation. It was trained on 38,693 lesions derived from 3D CT scans covering the entire body. However, it does not function as a traditional promptable model, as it operates on a fixed region of interest (ROI) with the expected lesion centered for segmentation. Consequently, it can only be utilized as a point model by employing center points as inputs.\\ 

\noindent In addition to these models, other notable promptable models exist, including 3D Sam Adapter \citep{3dsamadapter} and Prism \citep{prism}. However, these models operate in a closed-set manner, having been trained exclusively on specific datasets without the capability to segment arbitrary prompted classes. Therefore, they were excluded from our evaluations.

\section{Evaluation}
\label{sec:metrics_appendix}
For evaluating segmentation and tracking performance, we employ a range of metrics that capture both accuracy and robustness in handling diverse lesion sizes and positions. The single-timepoint segmentation models are evaluated on six held-out lesion segmentation datasets, as detailed in Table \ref{tab:test_datasets}. This dataset collection includes a multi-timepoint, in-house annotated whole-body melanoma dataset, which we use for all tracking model experiments through 5-fold cross-validation, addressing the lack of suitable public longitudinal datasets.

\begin{table*}[!ht]
    \centering
    \caption{\textbf{Test Datasets.}  This table summarizes the 6 datasets used for evaluating our proposed model. The datasets encompass a large set of annotated images from various types of lesions and institutions. For each dataset, we provide the name, the number of images, the specific types of lesions targeted, and links to access the datasets.}
    \label{tab:test_datasets}
    \resizebox{\textwidth}{!}{ 
    \begin{tabular}{lclr}
        \toprule
        Name & Images & Target & Link \\
        \midrule
        Liver Metastases & 171 & Colorectal Cancer & \url{www.cancerimagingarchive.net/collection/colorectal-liver-metastases/} \\ 
        Adrenal-ACC-Ki67-Seg & 53 & Adrenocortical Carcinoma & \url{www.cancerimagingarchive.net/collection/adrenal-acc-ki67-seg/} \\ 
        HCC-TACE-Seg & 66 & Primary Liver Cancer & \url{www.cancerimagingarchive.net/collection/hcc-tace-seg/} \\ 
        Lnq2023 & 393 & Malignant Lymph Nodes & \url{lnq2023.grand-challenge.org/} \\ 
        RIDER Lung CT & 55 & Lung Cancer & \url{www.cancerimagingarchive.net/collection/rider-lung-ct/} \\ 
        Whole-body Melanoma & 159 & Metastatic Melanoma & Private \\ 
        \bottomrule
    \end{tabular}
    }
\end{table*}

\begin{table*}[ht!]
\centering
\setlength{\tabcolsep}{4pt} 
\begin{tabular}{llccccccc|c}
\toprule
Dim & Model & Prompt & \shortstack{Colorectal\\ Liver Tumor} & \shortstack{Adrenal\\Tumor} & \shortstack{Primary\\Liver Cancer} & \shortstack{Lymph Node\\Metastases} & \shortstack{Lung\\Tumor}	 & \shortstack{Whole-body\\Melanoma}  & \shortstack{Avg.\\Dice}\\
\midrule 
\multirow{3}{*}{3D} & nnUNet~\cite{isensee_nnu-net_2021} & - & \colorrange{64.09} & \colorrangea{89.03} & \colorrangeb{72.27} & \colorrangec{43.34} & \colorranged{72.05} & \colorrangee{62.01} & 67.13\\ \addlinespace  \cline{2-10} \addlinespace
                  & \multirow{2}{*}{\textit{\textbf{Ours}}} & point & \colorrange{75.38} & \colorrangea{89.10} & \colorrangeb{78.39} & \colorrangec{75.60} & \colorranged{77.18} & \colorrangee{82.58} & \textbf{79.71}\\
                    &   & box & \colorrange{74.05} & \colorrangea{92.04} & \colorrangeb{85.71} & \colorrangec{80.63} & \colorranged{82.51} & \colorrangee{84.66} & \textbf{83.26}\\
\midrule
 \multicolumn{3}{r}{Dice Inter-Rater Variability} & 76 \cite{liver_interrater} & - & 84 \cite{liver_interrater} & 80 \cite{lymph_interrater} & 81-85 \cite{lung_interrater} & 80-85\cite{hering2024improving}\\
\bottomrule
\end{tabular}
\caption{\textbf{Performance Comparison Against Supervised Segmentation.} This table compares the segmentation performance of our model with nnUNet, a leading supervised medical segmentation model trained specifically on each test dataset and evaluated via 5-fold cross-validation. Remarkably, our model, despite never being trained on these held-out datasets, achieves higher Dice scores across all lesion types by leveraging either point or box prompts. Results are benchmarked against human inter-observer variability, offering an upper bound reference.}
\label{tab:results_st_supervised}
\end{table*}

\subsection{Single-Timepoint Segmentation Metrics}  
We evaluate segmentation performance using two primary metrics: 
\begin{itemize}
    \item \textbf{Dice Score}: Measures the overlap between the predicted and ground truth lesion masks.
    \item \textbf{Normalized Surface Dice (NSD)} with a 2mm tolerance: Ensures precise boundary delineation, accounting for varying lesion sizes by focusing on surface-level deviations. This metric is particularly suited for handling both small and large lesions.
\end{itemize}

\subsection{Tracking Metrics}  
For tracking across longitudinal scans, we prompt the previous timepoint and evaluate the model's performance on the follow-up segmentation using the following metrics:\\

\begin{itemize}
    \item \textbf{Center Point Matching (CPM@25)}: The percentage of ground truth and predicted lesion center points within a 25mm distance, reflecting the accuracy of lesion localization over time. We follow the 25mm threshold used in related work for consistency.
    \item \textbf{Dice@25}: The Dice score for lesions with center points correctly matched within 25mm, capturing segmentation quality for accurately tracked lesions.
    \item \textbf{Mean Euclidean Distance (MED)}: The average Euclidean distance between predicted and ground truth lesion center points, providing a direct measure of tracking precision. Lesions without a corresponding match in the ground truth or prediction are excluded from this calculation.
    \item \textbf{Total Dice Score}: The overall Dice score across all tracked lesions, assessing the model's ability to maintain segmentation quality over time, including missed or wrongly matched lesions.
\end{itemize}

\noindent All tracking metrics are averaged by patient. First, the average over all lesions of a particular scan is calculated, and then weighted by the number of scans per patient to account for variability in the number of available scans per patient.

\begin{figure*}[t]
  \centering
  \includegraphics[width=0.85\linewidth]{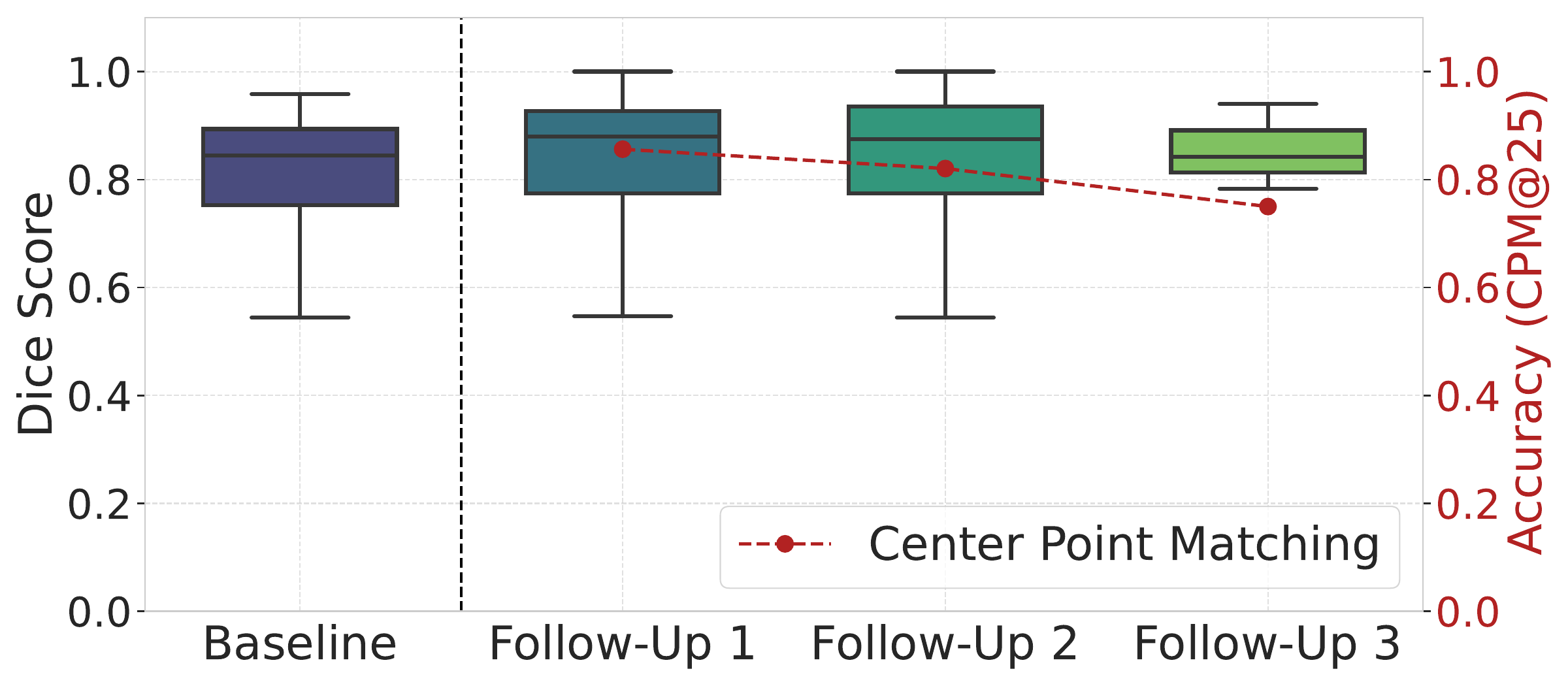}
  \caption{
\textbf{Consistent Lesion Tracking Performance Over Time Using Point Prompts.} Similar to Fig. \ref{fig:tracking_autoreg}, we show the initial Dice score distribution for the baseline scan using LesionLocator's segmentation model with \textit{point prompts}. For follow-up scans, tracking is performed autoregressively, as proposed, using prior masks as prompts. Tracking accuracy relative to the baseline is measured as CPM@25 (lesion matches within 25mm) with corresponding Dice@25 (Dice score of matched lesions). Similar to using box prompts in the first image, the Dice for matched lesions remains consistently high, with matching accuracy above 80\% and only a slight decrease over time. Note: Only a single patient in the dataset has a Follow-Up 3 scan, so this distribution is based on one scan with 4 lesions, of which 3 were correctly matched.}
  \label{fig:tracking_autoreg_point}
\end{figure*}

\section{Additional Results}
\label{sec:results_appendix}

\textbf{Zero-Shot Segmentation Performance Exceeds Supervised Models.} We benchmarked our zero-shot promptable segmentation model against nnUNet~\cite{isensee_nnu-net_2021}, a leading supervised segmentation framework that has consistently set high standards in medical image segmentation~\cite{nnunet_revisited}. To establish a robust comparison, nnUNet was trained independently on each of our six benchmark datasets, ensuring it had full access to the specific lesion types and image distributions within each dataset (see Tab. \ref{tab:results_st_supervised}). Remarkably, despite nnUNet’s access to the dataset from each specific lesion type, our zero-shot model outperformed it by over 15 Dice points on average, a significant margin that underscores the versatility and generalization capabilities of our approach.\\

\noindent Our model achieved these results without any prior exposure to the datasets or lesion-specific information, relying solely on prompts such as points or bounding boxes to localize regions of interest. This not only highlights the model’s zero-shot proficiency but also its robustness across varied anatomical contexts, from colorectal liver tumors to whole-body melanoma. In addition, our model’s performance in zero-shot settings closely approaches or even reaches inter-rater variability levels reported in literature, further reinforcing its reliability and potential as a scalable solution in clinical scenarios where labeled data may be limited or unavailable.\\

\noindent \textbf{Consistently High Tracking Performance Irrespective Of Prompt.} LesionLocator achieves robust and consistent temporal tracking accuracy, as demonstrated in Fig.~\ref{fig:tracking_autoreg} of the main paper. In this figure, the initial Dice distribution on the baseline scan shows strong performance using box prompts in the segmentation module. To complement this, Appendix Fig.~\ref{fig:tracking_autoreg_point} illustrates a similar initial distribution with less informative point prompts, which perform slightly lower overall but still demonstrate high accuracy. The bar plots for subsequent timepoints show that LesionLocator consistently achieves high Dice scores using autoregressive mask prompts, even when trained exclusively on consecutive image pairs. This highlights the generalizability of our longitudinal training approach. With minimal performance degradation across multiple timepoints, LesionLocator excels in lesion matching (CPM@25), ensuring reliable and sustained tracking throughout a sequence of scans irrespective of initial prompt type.\\

\noindent \textbf{Robustness.} To further evaluate out-of-distribution performance, we extended our experiments to assess robustness under real-world conditions. Specifically, our clinicians additionally annotated patients with diffuse, challenging-to-segment lesions from two centers. These cases feature varying resolutions, implant artifacts and were unseen during training. The results shown in Tab. \ref{tab:robustness_results} demonstrate that our method, despite the expected decrease, maintains robust performance. \\

\begin{table}[t]
\centering
\begin{adjustbox}{width=\columnwidth}
\setlength{\tabcolsep}{4pt} 
\begin{tabular}{l|lcccc}
\toprule
\textbf{Data} &\textbf{Model} & \textbf{Dice@25}$\uparrow$ & \textbf{MED}$\downarrow$\\
\multirow{3}{*}{\rotatebox[origin=c]{90}{\shortstack{unseen\\OOD}}} & Yan et.al~\cite{sam_tracking} \textit{(point tracker)} & - & 9.07 \\
  & Hering et.al~\cite{hering_tracking}(\textit{best baseline}) & 64.42 & 6.39 \\
  & \textbf{\textit{Ours (LesionLocator)}} & \textbf{76.55} & \textbf{5.13} \\
\bottomrule
\end{tabular}
\end{adjustbox}
\caption{Robust performance on diffuse-lesion diverse test set.}
\label{tab:robustness_results}
\end{table}